%% file: main.tex
\PassOptionsToPackage{dvipsnames}{xcolor} 

\newif\ifarxiv    
\arxivtrue 

\documentclass[11pt]{article}

\ifarxiv
    \usepackage[final]{acl}
\else
    \usepackage[review]{acl}
\fi

\usepackage{times}
\usepackage{latexsym}

\usepackage[T1]{fontenc}

\usepackage[utf8]{inputenc}

\usepackage{microtype}

\usepackage{inconsolata}

\usepackage{graphicx}

\usepackage[most]{tcolorbox} 
\usepackage{caption}         
\tcbuselibrary{breakable}

\usepackage{abbrev}

\usepackage{array}
\usepackage[dvipsnames]{xcolor}
\usepackage{tcolorbox}

\newcommand{\smallTitle}[1]{{\noindent\textbf{{#1}}}}

\usepackage[usestackEOL]{stackengine}

\usepackage{mathtools}

\newcommand{\defeq}{\coloneqq}

\newcommand{\E}{\mathbb{E}}

\newcommand{\Eb}[2]{\E_{#1}\!\left[#2\right]}

\newcommand{\bI}{\mathbf{I}}

\newcommand{\bzero}{\mathbf{0}}

\newcommand{\bx}{\mathbf{x}}

\newcommand{\bepsilon}{{\boldsymbol{\epsilon}}}

\newcommand{\bc}{\mathbf{c}}

\usepackage{listings}
\usepackage{xcolor}  

\definecolor{mylightgray}{gray}{0.95}

\lstdefinestyle{grayblock}{
  backgroundcolor=\color{mylightgray},
  basicstyle=\ttfamily\small,
  frame=single,
  breaklines=true,
  postbreak=\mbox{\textcolor{red}{$\hookrightarrow$}\space},
  tabsize=2,
  showstringspaces=false
}

\usepackage{algorithm}
\usepackage{algpseudocode}
\usepackage{float}

\usepackage{multirow, booktabs}
\usepackage{siunitx}
\usepackage{makecell}

\usepackage{booktabs}       
\usepackage{amsfonts}       
\usepackage{nicefrac}       
\usepackage{microtype}      
\usepackage{array,colortbl}
\usepackage{xcolor}
\usepackage{graphbox}
\usepackage{placeins}
\usepackage{wrapfig}
\usepackage{subcaption}
\usepackage{etoolbox}

\newlength\myindent
\usepackage{marvosym, ifsym}

\usepackage{graphicx}
\usepackage{amsmath}
\usepackage{booktabs}

\tcbset{
    mydoublebox/.style={
        colback=white,     
        colframe=mygray,        
        fonttitle=\bfseries,    
        coltitle=white,         
        boxrule=0.5pt,          
        width=0.9\textwidth,       
        top=2mm, bottom=2mm,    
        left=2mm, right=2mm,   
        enhanced,                
        breakable,
        unbreakable
    }
}

\usepackage{enumitem}

\usepackage{inconsolata}

\newcommand{\blfootnote}[1]{%
  \begingroup
  \renewcommand\thefootnote{}%
  \footnote{\scriptsize#1}%
  \addtocounter{footnote}{-1}%
  \endgroup
}

\usepackage{balance}

\title{VChain: Chain-of-Visual-Thought for Reasoning in Video Generation}

\ifarxiv
    \author{%
    Ziqi Huang, Ning Yu\textsuperscript{\Letter}$^{\dag}$, Gordon Chen, Haonan Qiu, Paul Debevec, Ziwei Liu\textsuperscript{\Letter} \\
      \\\vspace{15pt}
        \tt\normalsize\color{Magenta}\url{https://eyeline-labs.github.io/VChain}
    }
\fi

\begin{document}

\twocolumn[{%
           \renewcommand\twocolumn[1][]{#1}%
            \maketitle
            \begin{center}
                \centering
                \vspace{-50pt}
                \includegraphics[width=0.99\textwidth]{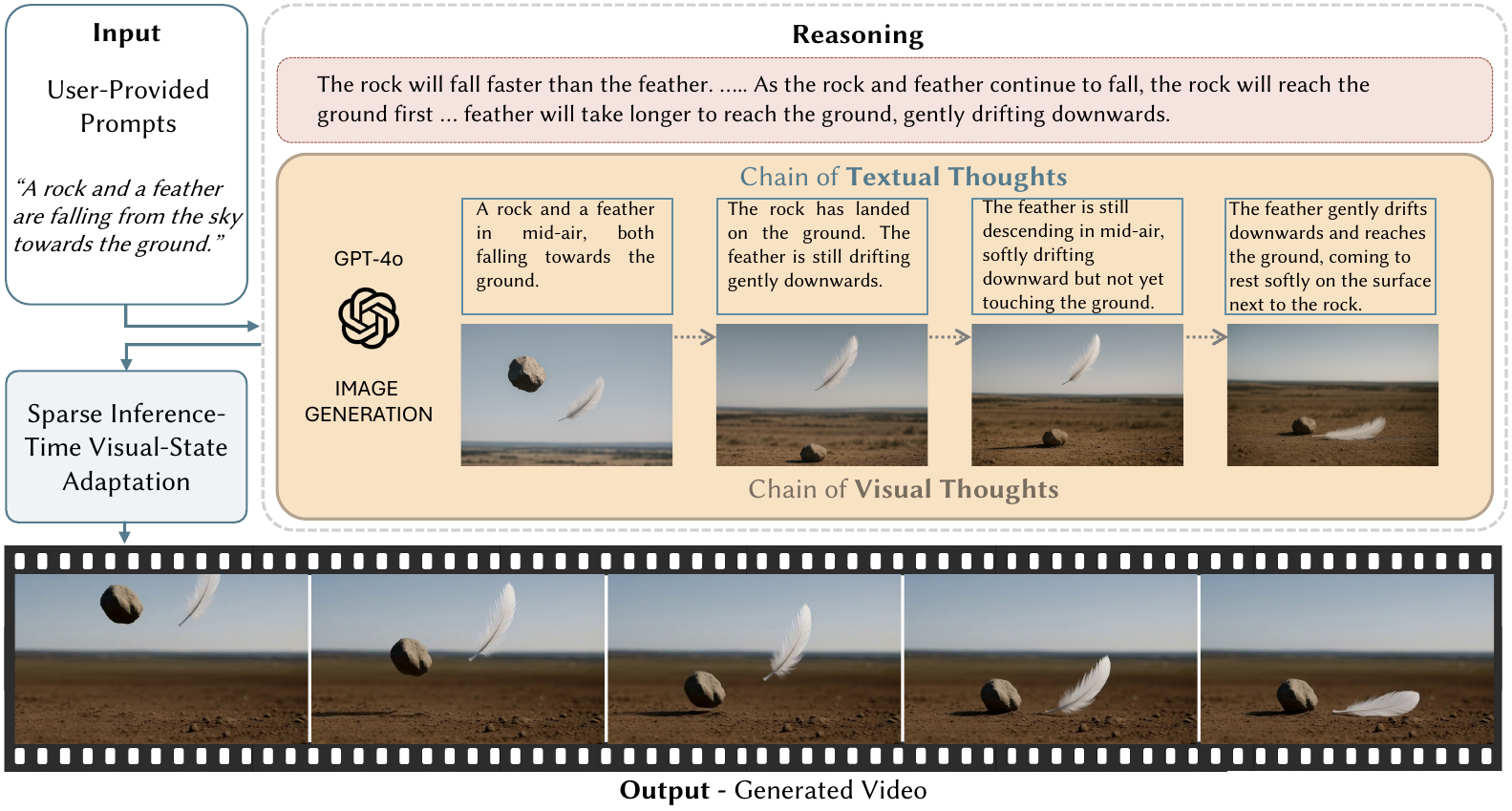}
                \captionof{figure}{\textbf{Overview of VChain.} We introduce \textbf{VChain}, an inference-time tuning framework for reasoning in video generation. Given a user-provided prompt (\textit{e.g.}, \textit{“A rock and a feather are falling from the sky towards the ground.”}), VChain leverages large multimodal models to generate a \textit{Chain of Visual Thoughts}, which are a sparse set of causally important keyframes to guide the video generator via \textit{Sparse Inference-Time Visual-State Adaptation}. VChain effectively improves reasoning in video generation without extensive re-training.
                }
                \vspace{-5pt}
                \label{fig:fig_paper_teaser}
                \vspace{10pt}
            \end{center}%
        }]


\ifarxiv
    \blfootnote{%
      \begin{itemize}[leftmargin=1.2em,itemsep=0.2ex,topsep=0ex,parsep=0pt,partopsep=0pt]
        \item \textsuperscript{\Letter}Corresponding Authors. $^{\dag}$Project Lead.
        \item Ziqi Huang, Gordon Chen, Haonan Qiu, and Ziwei Liu are with \textit{Nanyang Technological University}.
        \item Ning Yu and Paul Debevec are with \textit{Eyeline Labs}.
      \end{itemize}
    }
\fi
    
\input{sec/0_abstract}
\input{sec/1_intro}

\input{sec/2_relatedwork}

\input{sec/3_method}

\input{sec/4_experiments}

\ifarxiv
    \input{sec/6_limitations}

\input{sec/5_conclusion}
\fi

\ifarxiv    
    \input{sec/acknowledgement}
\fi
\clearpage

\bibliography{main}

\clearpage

\renewcommand\thesection{\Alph{section}}
\setcounter{section}{0} 

\input{sec/7_supplementarymaterials}

\clearpage

\end{document}

%% file: sec/0_abstract.tex
\begin{abstract}
Recent video generation models can produce smooth and visually appealing clips, but they often struggle to synthesize complex dynamics with a coherent chain of consequences. Accurately modeling visual outcomes and state transitions over time remains a core challenge. In contrast, large language and multimodal models (\eg, GPT-4o) exhibit strong visual state reasoning and future prediction capabilities. To bridge these strengths, we introduce \textbf{VChain}, a novel inference-time chain-of-visual-thought framework that injects visual reasoning signals from multimodal models into video generation. Specifically, VChain contains a dedicated pipeline that leverages large multimodal models to generate a sparse set of critical keyframes as snapshots, which are then used to guide the \textit{sparse inference-time visual-state adaptation} of a pre-trained video generator only at these key moments. Our approach is tuning-efficient, introduces minimal overhead and avoids dense supervision. Extensive experiments on complex, multi-step scenarios show that VChain significantly enhances the quality of generated videos.
\end{abstract}

%% file: sec/1_intro.tex
\section{Introduction}
\label{sec:intro}

Video generation~\cite{yang2024cogvideox, Gen3, kling, kong2024hunyuanvideo, wan2025wan, ma2025stepvideot2vtechnicalreportpractice, agarwal2025cosmos, Minmax} aims to synthesize coherent and realistic visual sequences, either from scratch or based on user-provided inputs such as text prompts, reference images, motion cues, or other forms of control. 
In recent years, this field has made remarkable progress, driven by powerful generative models such as diffusion models~\cite{sohl2015deep, song2020score, ho2020ddpm}, and supported by large-scale video datasets and increasing computational resources.

Modern video generation models have achieved impressive results in generating smooth and visually appealing video clips. However, they still struggle to reflect the intrinsic dynamics of the real world, especially when it comes to generating sequences that involve meaningful state transitions or coherent chains of consequences. As a result, current methods often fail to capture how visual states evolve over time in a logically consistent and causally grounded manner. For example, given a prompt like ``a person drops a cup, it hits the ground, and the liquid splashes out,'' many models may render smooth deformations between frames but omit key causal steps, such as the cup deforming on impact or the splash propagating outward, resulting in scenes that are logically inconsistent or physically implausible.

In contrast, large language and multimodal models excel precisely in the areas where video generation models tend to struggle. Models such as GPT-4o~\cite{hurst2024gpt4o} have made rapid progress in general reasoning and cross-modal understanding. These models show strong capabilities in following instructions, multi-step reasoning, and aligning semantics across text and vision. Although they do not explicitly simulate visual dynamics over time, they are effective in inferring likely transitions between visual states. For instance, they can reason that if a glass tips over, it may shatter, or that if a person jumps, they will eventually land. This ability to suggest causally and logically consistent progressions offers a promising signal that current video generators lack. A natural question is raised: can we leverage this reasoning ability from large multimodal models to guide video generation models towards more coherent chains of visual consequences?

To this end, we propose \textbf{VChain}, a novel inference-time tuning framework that introduces high-level reasoning into video generation. The core idea is to represent the evolution of a scenario as a sparse sequence of \textit{Visual Thoughts} - keyframes that capture critical intermediate states that a reasoning agent might anticipate.  These visual thoughts are automatically generated using large multimodal models and serve as guidance signals for the video generator.
VChain mainly consists of two main components. \textit{1) Visual Thought Reasoning:} We design a dedicated pipeline that leverages large multimodal models to decompose a user-provided text prompt into a concise set of causally important \textit{Visual Thoughts}. These keyframes capture the intended chain of visual outcomes and act as a blueprint for the temporal structure of the video. \textit{2) Sparse Inference-Time Visual-State Adaptation:} Then, the pre-trained video generator is quickly and efficiently fine-tuned using only the \textit{Visual Thought} keyframes. The model is adjusted in a focused manner at these critical visual states, allowing it to capture the intended visual state transitions. Compared to tuning on video data, this approach is significantly faster and more practical for deployment.

As an inference-time tuning method, VChain offers several benefits. 
\textit{(1) Self-contained:} All supervision is synthesized on the fly during inference by prompting a large multimodal model, with no need for external annotations, curated datasets, or retrieval systems.
\textit{(2) Efficient:} The tuning is only supervised by a few keyframes with limited iterations, and thus introduces minimal overhead relative to the cost of sampling the video itself.
\textit{(3) Effective:} We evaluate VChain on complex, multi-step video generation tasks that require strong causal reasoning. Across these scenarios, VChain consistently improves the dynamic fidelity of generated videos, leading to sequences that better reflect logical consequences, smooth transitions, and coherent visual narratives.
Beyond a specific technique, VChain offers a new pathway: treating multimodal models as reasoning modules that complement, rather than replace, generative models in constructing causally coherent visual narratives.
More generally speaking, VChain encourages the community to rethink how reasoning can be integrated into video generation - not through model retraining or dense supervision, but by transforming general-purpose multimodal intelligence into chain-of-visual-thought guidance at inference time.

In summary, our contributions are:
\begin{itemize}
\setlength{\itemsep}{0pt}
\setlength{\parskip}{0pt}
\setlength{\parsep}{0pt}
\item We introduce \textbf{VChain}, a novel framework that uses chain-of-visual-thought from large multimodal models to bring high-level reasoning into video generation.
\item We design the \textit{Visual Thought Reasoning} pipeline, a GPT-guided pipeline that synthesizes sparse, causally grounded keyframes for guiding video generation.
\item Extensive experiments demonstrate that sparse supervision on these keyframes improves a model’s ability to produce videos with coherent visual consequences and interpretable state transitions.
\item Our method operates entirely at inference time, requires no external training data, and adds minimal computational overhead.
\end{itemize}

%% file: sec/2_relatedwork.tex
\section{Related Work}
\label{sec:related_work}

\subsection{Video Generation}

Video generation has seen rapid progress~\cite{yang2024cogvideox, Gen3, kling, kong2024hunyuanvideo, wan2025wan, ma2025stepvideot2vtechnicalreportpractice, agarwal2025cosmos, Minmax, wang2025lavie, fan2025vchitect, zhang2023show1}, driven by advances in diffusion models~\cite{sohl2015deep, song2020score, ho2020ddpm, song2020ddim,  blattmann2023stable, esser2024scaling, ding2021cogview, ding2022cogview2, imagenvideo, zhang2023controlnet, mou2023t2iadapter, huang2023collaborative}, variational autoencoder-based compression~\cite{kingma2013vae, van2017vqvae, esser2021vqgan, podell2023sdxl, yu2023magvit}, and transformer-based backbones~\cite{dosovitskiy2020image, peebles2022scalable}.
Despite impressive progress in visual fidelity, smooth motion, and temporal alignment, most existing video generation methods remain limited to surface-level coherence~\cite{zheng2025vbench2, yue2025simulating}. They typically fail to capture deeper aspects such as causal dynamics, physical interactions, and meaningful state transitions. These models often overlook how actions lead to consequences, how objects behave under physical laws, or how scene states evolve with internal logic. To address this gap, we introduce an inference-time reasoning framework that injects high-level semantic supervision obtained from large multimodal models into the generation process. This approach enables pretrained video generators to produce outputs that are not only visually plausible but also causally and physically grounded.

\subsection{Multimodal Models for Understanding and Generation}

Large language models (LLMs) like GPT-4~\cite{openai2023gpt4} and multimodal models such as Gemini~\cite{team2023gemini} and GPT-4o~\cite{hurst2024gpt4o} have shown strong capabilities in vision-language tasks, including instruction-following, visual question answering, and interactive reasoning. These models can perform reasoning about visual scenes, and more importantly, understand and generate grounded visual content (See more discussions in \textit{Appendices}).
Recent works such as Transfusion~\cite{zhou2024transfusion} incorporate these capabilities into multimodal pipelines for image generation. LMD~\cite{lian2023lmd} and LVD~\cite{lian2023lvd} leverage a large language model to generate coarse layouts to guide visual synthesis. However, these approaches typically treat (M)LLMs as static prompt interpreters or high-level planners~\cite{li2023gligen,qu2023layoutllm, wang2026mavis}, or alternatively require dense retraining and architectural modifications~\cite{pan2025transfer, lin2025exploring}.
Unlike existing methods, we propose a lightweight inference-time reasoning framework for video generation that leverages off-the-shelf multimodal models and pre-trained video generators. Our method avoids dense retraining and instead injects high-level reasoning signals through sparse visual supervision, enabling more causally consistent and semantically grounded video generation with minimal overhead.

%% file: sec/3_method.tex
\section{The VChain Framework}

VChain is an inference-time reasoning framework designed to enhance the causal and physical coherence of video generation. Built on top of a pre-trained video generator, it aims to improve the model’s ability to reflect reasoning, physics, causality, and commonsense understanding, producing videos that are more physically grounded and causally consistent.

As shown in Figure~\ref{fig:fig_paper_framework}, the VChain framework has three key stages:
\textit{(1) Visual Thought Reasoning}, which uses a large multimodal model to infer key events and their consequences as a sparse sequence of visual snapshots;
\textit{(2) Sparse Inference-Time Visual-State Adaptation}, which injects these Visual Thoughts from stage 1 into the pre-trained video generator via lightweight LoRA adaptation; and
\textit{(3) Video Sampling}, which produces the final video by using both the stage-1 thoughts and the stage-2 tuned generator.

\subsection{Preliminaries}

\smallTitle{Diffusion Models.} 
Diffusion models are a class of generative models that reconstructs data $\bx_0$ such as natural images or videos by iteratively denoising starting from the Gaussian prior $\bx_T$. 
A widely used training loss~\cite{ho2020ddpm} is $L_\mathrm{DM}(\theta) \defeq \Eb{t, \bx_0, \bepsilon}{ \left\| \bepsilon - \bepsilon_\theta(\bx_t, t) \right\|^2}$, where $\bx_t$ is a noisy image or video obtained by adding noise $\bepsilon \sim \mathcal{N}(\bzero, \bI)$ to the original visual $\bx_0$. The network $\bepsilon_\theta(\cdot)$ learns to estimate this added noise.
To generate new data $\bx_0$, the trained model $\bepsilon_\theta(\cdot)$ denoises $\bx_t$ iteratively from $t = T$ to $t = 0$, using the predicted noise at each step.

\smallTitle{Video Diffusion Models.} Our work builds on Wan~\cite{wan2025wan}, a state-of-the-art video generation foundation model trained on a mix of video and image datasets, supporting both video and image generation. 
Recent progress in diffusion-based video generation has been shifting from U-Net~\cite{ronneberger2015unet} architectures to Diffusion Transformers (DiTs)~\cite{peebles2022scalable} with Flow Matching~\cite{lipman2022flow}. Wan adopts this newer paradigm, a design now common in text-to-video (T2V) systems~\cite{kong2024hunyuanvideo}. DiTs offer scalability advantages, while Flow Matching enables faster and more stable training convergence.
Wan includes three main components: \textit{1) Wan-VAE}: a spatio-temporal variational autoencoder; the \textit{2) video diffusion transformer}, and the \textit{3) text encoder}.
Given a video $V \in \mathbb{R}^{(1+T) \times H \times W \times 3}$, Wan-VAE compresses it into VAE latent $x \in \mathbb{R}^{(1 + T/4) \times H/8 \times W/8}$. The compression is spatial (by a factor of $8 \times 8$) for all frames, and temporal (by a factor of 4) for all frames except the first, which is only spatially compressed.
The Wan video generation model uses the flow matching~\cite{lipman2022flow,esser2024scaling} training objective in the Wan-VAE's latent space. Given the video (or image) latent $\bx_1$ and noise $\bx_0 \sim \mathcal{N}(\bzero, \bI)$, the noised latent is defined by the linear interpolation:
\begin{align}
    x_t &= tx_1 + (1 - t)x_0,
\end{align}
where the timestep $t \in [0, 1]$ is sampled from a logit-normal distribution. The model is trained to predict the velocity: 
\begin{align}
    v_t &= \frac{dx_t}{dt} = x_1 - x_0,
\end{align}
using the objective:
\begin{align}
    L_\mathrm{}(\theta)  &= \mathbb{E}_{\bx_0, \bx_1, \bc, t} \left\| u_\theta(\bx_t, t, \bc) - v_t \right\|^2,
    \label{eq:flow_matching_loss}
\end{align}
where $u_\theta$ is the denoising model and $\bc$ represents the embedded text prompt. The text encoder transforms the input text prompt into token embeddings, which we refer to as $\bc$ for brevity.

\smallTitle{Low-Rank Adaptation (LoRA).} 
LoRA~\cite{hu2022lora} is a parameter-efficient fine-tuning technique. It freezes the original model weights, and injects trainable low-rank decomposition matrices into network layers, largely reducing the number of trainable parameters.
Specifically, for a pre-trained weight matrix $W_0 \in \mathbb{R}^{d \times k}$, LoRA reparametrizes the update as $W_0 + \Delta W = W_0 + BA$,
where $B \in \mathbb{R}^{d \times r}$, $A \in \mathbb{R}^{r \times k}$, and $r \ll \min(d, k)$. Only $A$ and $B$ are updated during training. Given an input $x$, the modified forward computation is $h = W_0 x + \Delta W x = W_0 x + BAx$.
Because of the low-rank property, LoRA offers both computational and memory efficiency, making it a strong fit for fine-tuning large video diffusion models.

\subsection{Visual Thought Reasoning}
\label{subsec:visual_thought_reasoning}

\begin{figure*}[t]
  \centering

  \includegraphics[width=\linewidth]{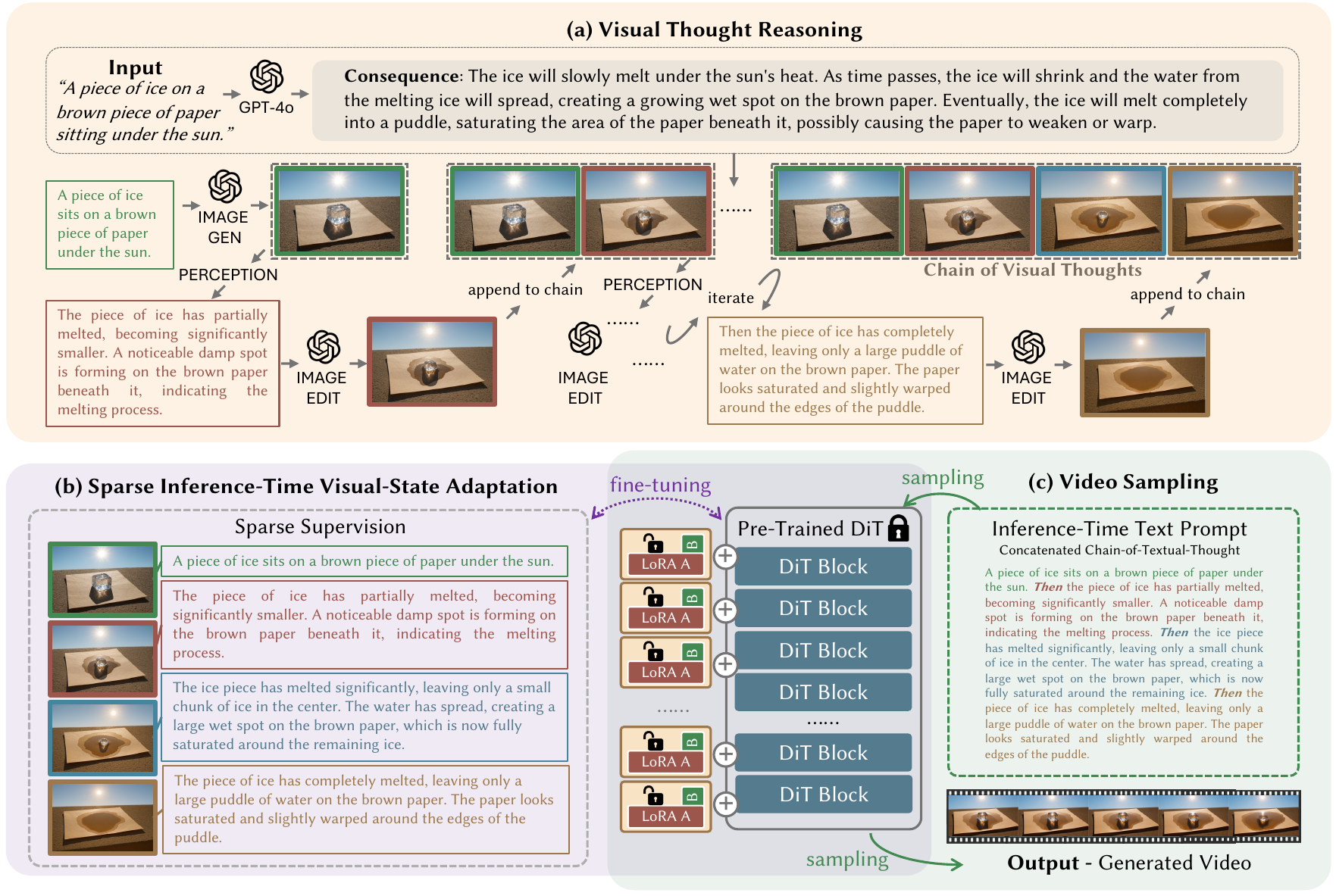}
  \vspace{-25pt}
  \caption{\textbf{VChain Framework.} An overview of our three-stage inference-time pipeline for reasoning in video generation. 
  \textbf{(a) Visual Thought Reasoning:} Given a user-provided text prompt, a large multimodal model (GPT-4o) infers a causal chain of events and generates a sequence of keyframes, termed the \textit{Chain of Visual Thoughts}, via iterative reasoning and image synthesis. 
  \textbf{(b) Sparse Inference-Time Visual-State Adaptation:} These visual thoughts (paired with their corresponding textual thoughts) serve as sparse supervision for fine-tuning a pre-trained video generator via LoRA. 
  \textbf{(c) Video Sampling:} The full sequence of textual thoughts is concatenated to form a single prompt, which is used to prompt the fine-tuned model in generating the final video output.}
  \label{fig:fig_paper_framework}
\end{figure*}

\definecolor{mylightgray}{gray}{0.6}  
\begin{algorithm}[t]
    \caption{Visual Thought Reasoning} \label{alg:vchain}
    {\footnotesize
    \begin{algorithmic}[1]
        \State given user-provided text prompt \textbf{p}

        \State
        \State {\color{mylightgray} \small{\texttt{\% generate first frame}}}
        \State $\textbf{txt}$, consequence = \texttt{chat}$(\textbf{p})$
        \State $\textbf{img}$ =  \texttt{image\_generate}$(\textbf{txt})$
        \State  {$\textbf{chain}_{vis} = [\textbf{img}]$
        \small{{\color{mylightgray} \texttt{\% init chain-of-visual-thought}}}}
        \State  {$\textbf{chain}_{txt} = [\textbf{txt}]$
        \small{{\color{mylightgray} \texttt{\% init chain-of-textual-thought}}}}
        
        \State
        \State {\color{mylightgray} \small{\texttt{\% iteratively generate subsequent frames}}}
        \State \textbf{repeat}         

             \State\indent $\textbf{txt}$, flag $=$ \texttt{perception}$(\textbf{chain}_{vis}$, consequence, $\textbf{p})$
             \State\indent $\textbf{img} =$ \texttt{image\_edit}$(\textbf{chain}_{vis}, \textbf{txt})$
             \State\indent $\textbf{chain}_{vis}$.\texttt{append}$(\textbf{img})$
             \State\indent $\textbf{chain}_{txt}$.\texttt{append}$(\textbf{txt})$

        \State \textbf{until} flag==terminate

        \State
        \State return $\textbf{chain}_{vis}, \textbf{chain}_{txt}$
        
    \end{algorithmic}
}
\end{algorithm}

%

Given a user-provided text prompt $\textbf{p}$ for video generation, we leverage the powerful multimodal reasoning capabilities of GPT-4o~\cite{hurst2024gpt4o} to generate a sequence of images, referred to as the \textit{Chain of Visual Thoughts}, that capture the key moments of the intended video. The steps and definitions of \textit{Visual Thought Reasoning} are listed in Algorithm~\ref{alg:vchain}.

We first prompt GPT-4o to reason about the likely outcome implied by the user input prompt $\mathbf{p}$. As illustrated in Figure~\ref{fig:fig_paper_framework}, given a prompt, “A piece of ice on a brown piece of paper sitting under the sun”, GPT-4o infers that the ice will melt due to the heat, forming a puddle that soaks the paper. This step establishes the ground-truth trajectory of the intended video, referred to as the \textit{consequence}, which serves as the basis for constructing the key transitions of the unfolding scene.

We then instruct GPT-4o to generate a caption $\mathbf{txt}_0$ describing the first frame in the \textit{Chain of Visual Thoughts}, which is transformed into an image $\mathbf{img}_{0}$ using GPT-4o's native image generation module. After that, GPT-4o predicts an editing instruction $\mathbf{txt}_i$ to produce the key moment at time step $i$ in our chain, conditioned on $\mathbf{p}$, the \textit{consequence}, and the \textit{Chain of Visual Thoughts} at the current timestep, $\textbf{chain}_{\text{vis}} =[\mathbf{img}_0, \mathbf{img}_1,\dots, \mathbf{img}_{i-1}]$. Then $\mathbf{txt}_i$ is used to generate the subsequent image $\mathbf{img}_{i}$. This process continues iteratively, where GPT-4o predicts an editing instruction and generates a corresponding image, and terminates only when the \textit{consequence} has been fully captured by $\textbf{chain}_{\text{vis}}$. 
 
The resulting output is a coherent sequence of keyframes, or \textit{Chain of Visual Thoughts} $ [\mathbf{img}_0$, $\mathbf{img}_1$,$\dots$, $\mathbf{img}_{N-1}]$ paired with its corresponding textual thoughts $ [\mathbf{txt}_0$, $\mathbf{txt}_1$, $\dots$, $\mathbf{txt}_{N-1}]$, that captures the temporal evolution implied by the user prompt. This approach also allows users to generate causally consistent image sequences without having to explicitly anticipate or specify the underlying consequences of the described scenario. Please refer to the \textit{Appendices} for detailed descriptions of the \textit{Visual Thought Reasoning} process, including system prompts, intermediate outputs, and workflow details.

\begin{table*}[t!]
  \caption{\textbf{Quantitative Evaluation.} VChain is compared with existing methods and ablation variants, achieving comparable or superior performance across all evaluation metrics.}
\label{tab:quantitative_evaluation}
  \resizebox{\textwidth}{!}{
  \begin{tabular}{c|c|ccc|ccc}
    \toprule
    \textbf{Method} & \textbf{\Centerstack{VBench\\Quality Score}} & \textbf{\Centerstack{Frame\\Quality}} & \textbf{\Centerstack{Temporal\\Smoothness}} & \textbf{\Centerstack{Video-Text\\Alignment}} & \textbf{\Centerstack{Physics}} & \textbf{\Centerstack{Commonsense\\Reasoning}} & \textbf{\Centerstack{Causal\\Reasoning}}   \\    \midrule
    T2V & 76.21\% & 57.24\% & 43.65\% & 40.04\% & 32.03\% & 32.42\% & 32.81\%  \\
    T2V + Prompt Aug& 77.51\% & 55.47\% & 50.59\% & 47.66\% & 38.09\% & 38.48\% & 41.99\% \\
    \hline
    Without Visual Thought & 78.47\% & 64.26\% & 52.93\% & 54.69\% & 44.14\% & 43.75\% & 47.51\% \\
    Without Sparse Tuning & 73.35\% & 44.07\% & 29.19\% & 42.97\% & 33.24\% & 34.57\% & 34.46\% \\
    \hline
    \textbf{VChain (Ours)} & \textbf{78.49\%} & \textbf{71.67\%} & \textbf{65.82\%} & \textbf{67.77\%} & \textbf{58.01\%} & \textbf{60.16\%} & \textbf{62.12\%} \\
  \bottomrule
\end{tabular}
}
\end{table*}

\subsection{Sparse Inference-Time Visual-State Adaptation}

Given the sparse and causally grounded \textit{Chain of Visual Thoughts} generated from the previous stage, we perform lightweight inference-time tuning on a pre-trained video generator.
We only use these keyframes as supervision, treating them as anchor points that encode important state changes (\eg, melting, breaking, or object movement).

Formally, let $\textbf{chain}_{\text{vis}}$ $=$ $[$$\mathbf{img}_0$, $\mathbf{img}_1$,$\dots$, $\mathbf{img}_{N-1}$$]$ be the sequence of $N$ \textit{Visual Thoughts} (keyframes), and $\textbf{chain}_{\text{txt}} = [\mathbf{txt}_0$, $\mathbf{txt}_1$, $\dots$, $\mathbf{txt}_{N-1}]$ be their corresponding \textit{Textual Thoughts}. Each $\mathbf{img}_i$ is treated as a one-frame video, paired with the caption $\mathbf{txt}_i$. These pairs $(\mathbf{img}_i, \mathbf{txt}_i)$ serve as the training data for tuning the video diffusion model using the same flow-matching objective as  Equation~\ref{eq:flow_matching_loss}:
\begin{align}
\mathcal{L}_{\text{vchain}}(\theta)
&= \mathbb{E}_{\bx_0, \bx_1, \bc, t}
\left\| u_\theta(\bx_t, t, \bc) - v_t \right\|^2 ,
\end{align}
where $\bx_1=\mathbf{img}_i$, $\bx_0 \sim \mathcal{N}(\mathbf{0}, \mathbf{I})$, $t \in [0, 1]$ is sampled from a logit-normal distribution, $\bx_t = t \bx_1 + (1 - t) \bx_0$ as in the flow-matching setup, and  $\bc$ is the text embedding of $\mathbf{txt}_i$.

This sparse tuning scheme offers two key benefits:
\textit{1) Focused supervision}: By concentrating only on keyframes that encode the critical moments (\eg, object breaking, melting, or appearing), we guide the model to focus on inferring causal outcomes and key visual state transitions.
\textit{2) Efficiency}: Since the tuning is \textit{image-only}, tuning is fast and memory-efficient. This makes our method practical for inference-time adaptation.
Additionally, our tuning does not require additional databases or labels. The entire supervision signal is generated internally from the \textit{Visual} and \textit{Textual Thoughts} (Section~\ref{subsec:visual_thought_reasoning}), making VChain easily pluggable into general pre-trained video generators.

\subsection{Video Sampling}

Following \textit{Sparse Inference-Time Visual-State Adaptation}, we concatenate every textual thought $\textbf{txt}_i$ from the \textit{Chain of Textual Thoughts} $\textbf{chain}_{txt}$ into a single composite prompt $\textbf{txt}_{concat}$. This final prompt is used as the input to the fine-tuned video generator to produce the output video. The resulting generation reflects both the inferred sequence of events and the adapted capabilities of the model.

%% file: sec/4_experiments.tex
\section{Experiments}

\subsection{Experimental Setup}

For \textit{Visual Thought Reasoning}, we use the GPT family~\cite{hurst2024gpt4o} as our large multimodal model. Specifically, we use \texttt{gpt-4o} for chat and perception, and \texttt{gpt-image-1} for steps involving image generation and editing. Our main experiments are conducted using the state-of-the-art pre-trained video generator Wan2.1-T2V-1.3B~\cite{wan2025wan}. We design 20 diverse test scenarios for both human evaluations and quantitative comparisons. We list the implementation details, test cases, and the time cost breakdown in \textit{Appendices}.

\subsection{Comparison Methods}
We compare our proposed method VChain against several baselines and ablation variants.

\smallTitle{Baseline Comparison.}
We include the following baselines:

\noindent\textit{- T2V:} The original pre-trained text-to-video generation model without any modification.

\noindent\textit{- T2V + Prompt Aug:} The input text prompt is enhanced using GPT-based prompt augmentation.

\smallTitle{Ablation Study.}
To further understand the impact of each component in VChain, we design the following ablation settings:

\noindent\textit{- Without Visual Thought:} We use our \textit{Visual Thought Reasoning} pipeline to produce both composite text prompts $\textbf{txt}_{concat}$ and visual thoughts $\textbf{chain}_{\text{vis}}$, but only feed $\textbf{txt}_{concat}$ to the video generator, omitting the visual thoughts for sparse tuning. This ablation evaluates the necessity of performing chain-of-thought reasoning \textit{visually}, showing that text-only thoughts are insufficient for reasoning in video generation.

\noindent\textit{- Without Sparse Tuning:} We use GPT-generated keyframes as-is for direct video interpolation, without fine-tuning the video generator. This variant evaluates the necessity of sparse tuning to align the dynamics with the inferred reasoning signals.

\noindent\textit{- VChain (Ours):} Our full framework, which combines both \textit{Visual Thought Reasoning} and \textit{Sparse Inference-Time Visual-State Adaptation} to enable reasoning in video generation.

\begin{figure*}[t]
  \centering
  \includegraphics[width=\linewidth]{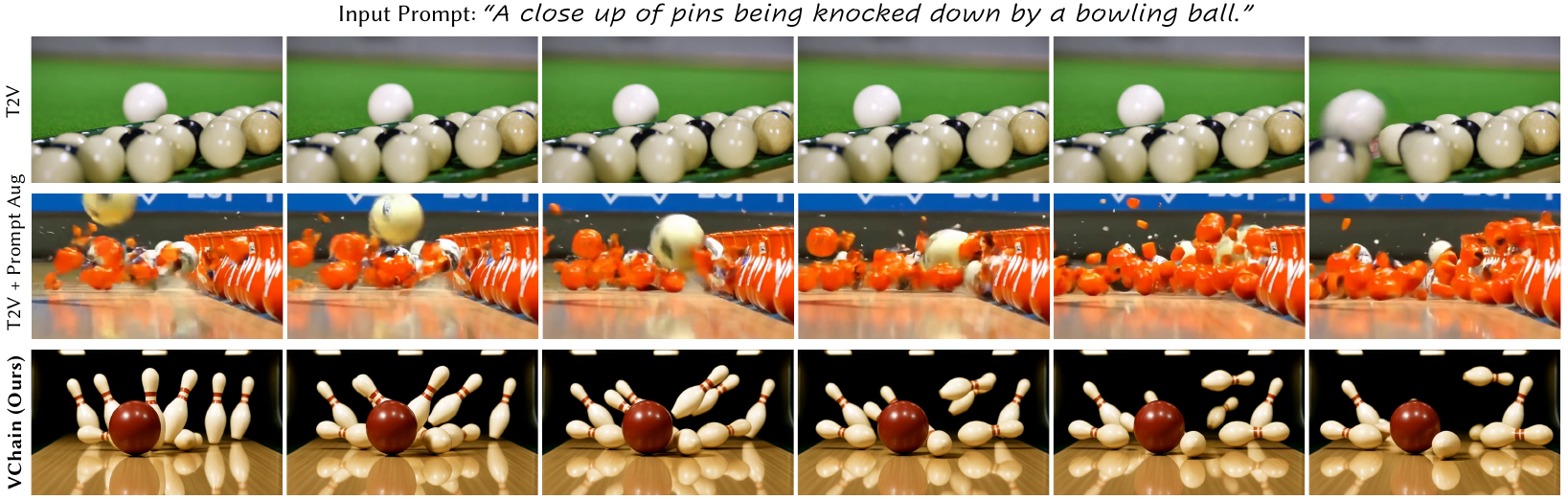}
  \caption{\textbf{Qualitative Results - Baseline Comparison.} 
\textit{T2V} fails to capture the key causal interaction: the pins remain mostly static or jitter slightly, with no meaningful collision, revealing a lack of physical reasoning despite temporal coherence.
\textit{T2V + Prompt Aug} introduces relevant elements and motion, but the dynamics are erratic and implausible. Pins deform unnaturally, visual artifacts appear, and later frames become unstable, indicating poor spatial consistency.
In contrast, \textit{VChain (Ours)} produces a coherent and physically realistic sequence: the ball strikes the pins with plausible force, leading to consistent knockdown. Object geometry and material properties are well preserved across frames. These results show that VChain not only enables causal reasoning about the outcome of physical interactions,  but also stabilizes spatial transitions.
}
  \label{fig:fig_paper_bowling}
\end{figure*}

\begin{figure*}[t]
  \centering
  \includegraphics[width=\linewidth]{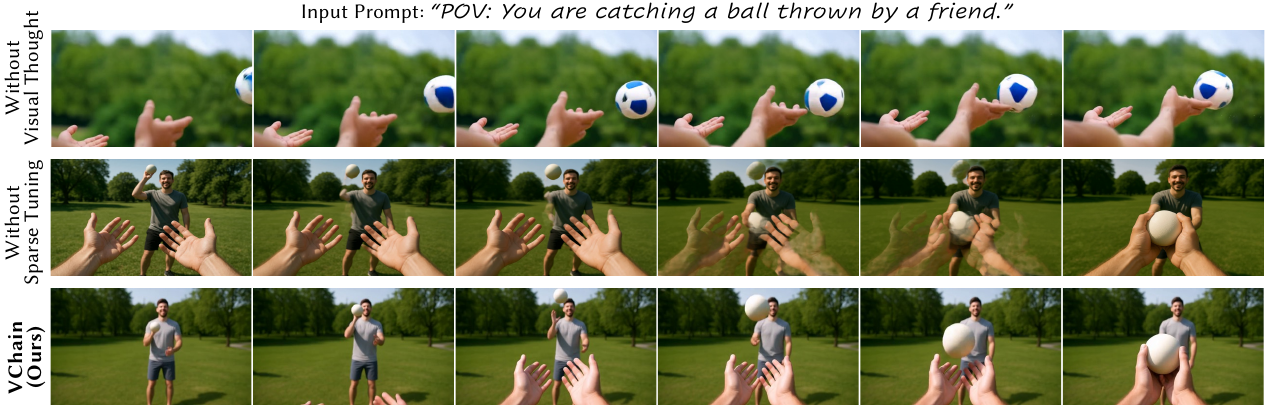}
  \vspace{-15pt}
    \caption{\textbf{Qualitative Results - Ablation Study.}     We compare VChain with two ablated variants. \textit{(1) Without Visual Thought:} Although the model recognizes that the video should be in a first-person perspective based on the textual prompt, it fails to capture the correct visual pattern for a ball-catching viewpoint. In contrast, VChain leverages the reasoned Visual Thoughts to render step-by-step intermediate visual states of the throw-and-catch process.  \textit{(2) Without Sparse Tuning:} While Visual Thoughts are included, the model performs direct frame interpolation without tuning, leading to warping artifacts due to spatial misalignments among individual frames in Visual Thoughts. VChain (Ours) produces the most coherent and physically grounded interaction, correctly depicting the ball being thrown and caught from a first-person perspective. Removing either component degrades video synthesis quality.
}
      \label{fig:fig_paper_catch_ball}
\end{figure*}

\subsection{Quantitative Comparisons}
\label{subsec:vbench_evaluation}
We introduce the evaluation metrics below.

\smallTitle{VBench Quality Score.}
To evaluate VChain's impact on fundamental video quality, independently of its reasoning or causal capabilities, we conduct quantitative evaluations using VBench~\cite{huang2024vbench, huang2024vbench++}, an evaluation framework designed to assess key technical dimensions of video generation, such as frame-level fidelity, temporal consistency, and motion dynamics \textit{etc}.
As shown in Table~\ref{tab:quantitative_evaluation}, VChain achieves comparable or slightly better scores than both the original pre-trained generator and other baselines.

We also perform complementary human evaluations focused on three core aspects of video quality:

\smallTitle{Frame Quality.} Visual quality of individual frames: aesthetics, sharpness, and realism.

\smallTitle{Temporal Quality.} Motion smoothness, temporal consistency, and dynamic realism across frames.

\smallTitle{Video-Text Alignment.} How faithfully the generated video reflects the input text prompt.

While VChain is primarily designed to enhance high-level reasoning in video generation (\textit{e.g.}, commonsense, causality, and physics), the results shown in Table~\ref{tab:quantitative_evaluation} confirm that it does not compromise basic visual quality. In fact, it often brings modest improvements. To directly assess VChain’s reasoning capabilities, we conduct targeted human studies along the following dimensions:

\smallTitle{Physics.} Evaluates whether the video follows physical laws, like gravity and air friction (\eg, rocks fall faster than feathers in the air). Participants rate how well the video obeys the laws of physics.

\smallTitle{Commonsense Reasoning.} Assesses whether events in the video reflect everyday real-world knowledge. For instance, blue paint mixed with yellow turning green, or oil floating on water. Users rate how well the video reflects common sense.

\smallTitle{Causal Reasoning.} Evaluates whether the video captures appropriate cause-and-effect relationships. Examples include a stone causing a splash when dropped in water, a ball failing to bounce on a pillow, or a switch turning on light. Participants are asked: ``How well does the video reflect the causal consequences of the initial setup?”

Human evaluators were presented with generated videos alongside their corresponding input prompts. The outputs from our method and the baselines were shown in randomized order to avoid bias. A total of 32 evaluators rated each video on a scale from 1 to 5 for each evaluation dimension. The scores are then averaged and normalized to a percentage scale, as reported in Table~\ref{tab:quantitative_evaluation}.

VChain consistently outperforms the baseline methods, particularly in reasoning-related dimensions such as physics, commonsense, and causality. These improvements demonstrate the effectiveness of the integration of our framework in inference-time reasoning for video generation.

\subsection{Qualitative Comparisons}

\ifarxiv
    Extensive qualitative results and comparisons are also provided in the \textit{Appendices}.
\fi

\smallTitle{Baseline Comparison.}
We present qualitative comparisons against baseline methods in Figure~\ref{fig:fig_paper_bowling}.
In the \textit{T2V} baseline, the model fails to produce any meaningful physical interaction: the pins remain mostly static or exhibit minor jittering, with no visible impact or knockdown. Although temporally stable, the output is semantically misaligned with the input prompt, lacking the key causal event of a bowling ball knocking down pins.
The \textit{T2V + Prompt Aug} variant introduces the ball and pins, showing some degree of collision and motion. However, the dynamics are chaotic and physically implausible. Pins deform or scatter in erratic ways, and the scene suffers from visual artifacts and temporal instability, particularly in later frames.
In contrast, \textit{VChain (Ours)} produces a coherent and physically grounded sequence. The bowling ball hits the pins with a realistic impact, and the pins fall in directions consistent with expected physical behavior. This outcome is enabled by chain-of-visual-thought reasoning, which provides the model with a structured, causal progression of events. Furthermore, object geometry and material properties are well preserved. Pins and the ball are visually distinct and accurately rendered.

\smallTitle{Ablation Study.}
In Figure~\ref{fig:fig_paper_catch_ball}, we compare VChain with two ablated variants: \textit{1) Without Visual Thought}, while it understands that the first-person perspective should be generated from the \textit{Textual Thoughts}, it fails to envision the correct visual pattern of a ball-catching POV. In contrast, our method benefits from directly ``seeing'' the Visual Thoughts, enabling accurate spatial understanding and rendering of the interaction. \textit{2) Without Sparse Tuning}, which includes Visual Thoughts directly performs frame interpolation, and warping artifacts emerge when attempting to bridge spatial misalignments between Visual Thought keyframes. \textit{VChain (Ours)} produces the most coherent and physically grounded interaction, accurately depicting the ball being thrown and caught. Removing either component leads to degraded video synthesis.

Figure~\ref{fig:fig_paper_comparison_5baselines}(b) in \textit{Appendices} highlights another example of a rubber duck and a rock falling into water. Without \textit{Visual Thought}, the duck appears submerged in water, violating the basic physical intuition that rubber ducks are supposed to float. In contrast, our method correctly depicts the duck floating on the water’s surface. This underscores the importance of having \textit{Visual Thoughts} (versus \textit{Textual Thoughts} only) at inference time: 
it’s important to view the \textit{Visual Thoughts} during inference - to actually ``see'' how the rubber duck floats on the water surface rather than sinks.
\ifarxiv
    Our demo video provides a more intuitive comparison.
\fi

%% file: sec/6_limitations.tex
\section{Limitations}

\subsection{Limitations of Visual Thought Generation}

Our framework inherits several limitations from GPT-4o. First, \texttt{gpt-image-1} tends to produce oversaturated and over-smooth images. Since frames are iteratively fed back as input, these artifacts accumulate, causing a yellow color cast and reduced photorealism in later frames (see \textit{Appendices}). Second, reliance on proprietary APIs introduces costs: each keyframe requires two calls, making total calls linear and token consumption quadratic. While this may limit accessibility for those with restricted budgets, the practical overhead remains modest as inference typically requires only 3--6 images. Third, while GPT-4o provides robust reasoning, its closed-source nature hinders customization and full reproducibility. As VChain is a general framework, integrating it with open-source MLLMs once they become sufficiently capable is highly valuable. Such an implementation will enhance accessibility and reproducibility for the broader research community.

\subsection{Limitations of Sparse Inference-Time Visual-State Adaptation}

Our method fine-tunes a pre-trained video generator using several keyframes, referred to as \textit{Visual Thoughts}, as supervision. This sparse tuning introduces an inherent trade-off: optimizing too strongly on static keyframes may reduce motion dynamics, since the model adapts primarily to still images, while insufficient optimization may weaken the reasoning signals injected into the generator, producing results closer to the untuned baseline.

Despite the potential trade-off, this \textit{sparse} tuning strategy offers two main advantages:
\textit{(1) Focused adaptation}: the model concentrates its capacity on semantically critical transitions (\eg, melting, breaking, or object interactions) rather than reconstructing entire video sequences.
\textit{(2) Efficiency}, as it eliminates the need for dense videos, significantly reducing both data preparation and computational overhead. This makes our approach well-suited for inference-time integration into existing pipelines.

Overall, while sparse supervision cannot fully capture video dynamics, the improvements in semantic alignment and causal coherence generally outweigh the loss in dynamics. This paradigm also challenges the conventional assumption that full video sequences are required for fine-tuning, showing that a carefully selected set of keyframes can provide sufficient guidance for adapting video generators to new prompts or scenarios.

\section{Ethical Considerations}

While both large multimodal models and video generators can produce vivid and compelling content, users should exercise caution when using AI-generated media. Outputs may inherit and amplify safety concerns and biases from the multimodal models and video generators they rely on. We strongly advocate for the responsible and ethical use of generative models.

\smallTitle{Potential Risks}. VChain is intended as a research contribution, but its ability to improve causal and physical coherence also increases the realism of synthetic videos. This realism could be misused for harmful purposes such as producing disinformation, deepfakes, or fabricated evidence. Moreover, because VChain depends on large multimodal models and pretrained generators, it might propagate their biases into more coherent video narratives, which may reinforce stereotypes or exclusion. We emphasize that VChain is designed for controlled research and creative exploration, not deployment in sensitive or adversarial settings.

%% file: sec/5_conclusion.tex
\section{Conclusion}
In this work, we present \textbf{VChain}, a general inference-time framework that integrates multimodal reasoning into video generation. By representing a scenario as a sparse sequence of \textit{Visual Thoughts} - keyframes capturing critical intermediate states inferred by large multimodal models - VChain injects causal and commonsense reasoning signals directly at inference time. This paradigm enables video generators to model meaningful state transitions without dense annotations or costly retraining.
Experiments on complex, multi-step scenarios show that VChain substantially improves the coherence, causal consistency, and rationality of generated videos, while maintaining efficiency and visual quality. More broadly, VChain demonstrates how the reasoning capabilities of large multimodal models can be effectively combined with the rendering and motion priors of video generators. We view this framework as a step toward bridging reasoning and generation, and hope to inspire further research on reasoning for video generation.

%% file: sec/acknowledgement.tex
\section*{Acknowledgments}
This study is supported by the Ministry of Education, Singapore, under its MOE AcRF Tier 2 (MOET2EP20221-0012, MOE-T2EP20223-0002). This research is also supported by cash and in-kind funding from NTU S-Lab and industry partner(s), and Eyeline Labs.

%% file: sec/7_supplementarymaterials.tex
\appendix

\section*{Appendices}

We provide additional implementation details in Appendix~\ref{supp_sec:implementation}, and qualitative results in Appendix~\ref{supp_sec:qualitative}.
Furthermore, we provide a dedicated discussion on the evolution of reasoning paradigms for video generation and how VChain fits into this taxonomy in Appendix~\ref{sec:reasoning_discussion}.
\ifarxiv
    A demo video is also available at \href{https://www.youtube.com/watch?v=HV4uAHJwt1k}{this link}.
\fi

\section{Additional Implementation Details}
\label{supp_sec:implementation}

\subsection{Implementation Details of Visual Thought Reasoning}
Given a user-provided input prompt describing a video, our \textit{Visual Thought Reasoning} pipeline synthesizes a sequence of keyframes which form the crucial moments of the video. The implementation details are as follows.

We first prompt GPT-4o's chat completions API with the system message shown in Figure~\ref{fig:first_frame_system_message} to instruct the model to reason about the video's likely spatial layout and anticipated causal consequences based on the user-provided input prompt. We employ LangChain \cite{langchain} to convert GPT-4o's unstructured textual outputs into structured schema-aligned responses containing:
\begin{enumerate}[itemsep=0pt, topsep=0pt, parsep=0pt, partopsep=0pt]
    \item \textit{Context Frame:} A richly detailed prompt used to generate the first frame in the \textit{Chain of Visual Thoughts}. 
    \item \textit{Concise Prompt:} A concise version of the Context Frame prompt (the full version is too long, so the first image is paired with this concise prompt during sparse inference-time visual-state adaptation).
    \item \textit{Consequences:} A sequence of inferred physical outcomes that define the expected trajectory of the generated video.
\end{enumerate}

The Context Frame is passed to GPT's \texttt{gpt-image-1} API to produce the corresponding image.

To generate subsequent keyframes in the \textit{Chain of Visual Thoughts}, we concatenate all previously generated images in our chain into a single composite image, as shown in Figure~\ref{fig:fig_paper_framework}.  This stitched chain of images, together with the user input prompt and the inferred consequences, is passed to GPT-4o's chat completion API using the system message in Figure~\ref{fig:next_frame_system_message}. GPT-4o predicts the next key moment in the sequence. Specifically, the output contains: 1) an image-editing instruction and 2) a boolean flag indicating whether a terminal state has been reached. We pass the same inputs as before along with the editing instruction to the \texttt{gpt-image-1} API to generate the next keyframe. We repeat this process iteratively, where we predict the next key moment and generate the corresponding image, until the boolean flag signals that the full sequence of consequences have been realized by the chain.

All outputs, including keyframe captions and reasoning chains, are stored in a structured JSON file (see Figure~\ref{fig:reasoning_output_example}).
We then generate a CSV file where each row contains an image file path and its corresponding caption, forming the image-text pairs used to fine-tune the video generation model. The first image is paired with the concise prompt, while each subsequent image is paired with its keyframe (example in Figure~\ref{fig:csv_output_example}).

\begin{figure*}[h!]
  \centering
  \includegraphics[width=\linewidth]{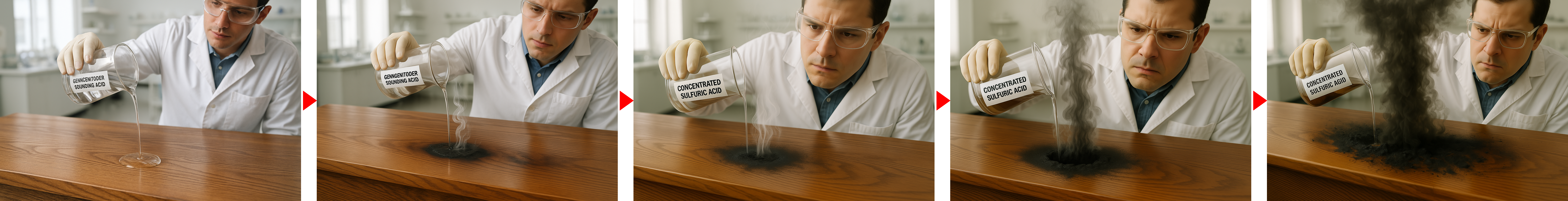}
  \caption{\textbf{Example of Visual Thoughts.} We show the reasoned Visual Thoughts of the input prompt: ``Concentrated sulfuric acid is poured onto a wooden table''. The sequence illustrates our pipeline’s inferred causal progression across keyframes.
  }
  \label{fig:img_sulfuric_acid}
\end{figure*}

Figure~\ref{fig:img_sulfuric_acid} shows an example of \textit{Chain of Visual Thoughts} generated by our pipeline.

\subsection{Time Cost}

Table~\ref{tab:time_cost} summarizes the average runtime of each stage in VChain, providing a detailed breakdown of the overall computational cost.

\subsection{Implementation Details of Sparse Inference-Time Visual-State Adaptation}

Our main experiments are conducted using the state-of-the-art pre-trained video generator Wan2.1-T2V-1.3B~\cite{wan2025wan}. We use the learning rate of $1e-4$, and fine-tune with a \texttt{train\_lora\_rank} of 16, and \texttt{train\_lora\_alpha} of 16.

\subsection{Details of Test Cases}

We design twenty test cases to support both human and quantitative evaluations.  
Each case depicts a simple, physically grounded scenario that requires causal reasoning to generate coherent outcomes.

{\itshape
\begin{itemize}[itemsep=0pt, parsep=0pt, topsep=0pt, partopsep=0pt]
    \item A rock and a feather falling from the sky towards the ground.
    \item An egg falling from the sky towards concrete ground.
    \item An ice cream cone is left out in the sun.
    \item A rubber duck and a rock fall into a water tank.
    \item A steel ball is dropped into water.
    \item Milk is poured into a cup of black coffee.
    \item A man falls off a pile of bricks.
    \item A steel ball falling through the air onto ice.
    \item A ball is dropped onto a pillow.
    \item A sandwich rotting over time.
    \item An elderly blows out a cake filled with candles.
    \item Red and yellow paint are mixed together with a brush.
    \item Concentrated sulfuric acid is poured onto a wooden table.
    \item An egg is dropped onto a pillow.
    \item A mailbox rusting over time in broad daylight.
    \item A man blows into a deflated balloon.
    \item Oil is poured into a glass of milk.
    \item A chameleon eats a flying insect.
    \item Blue and yellow paint are mixed together with a brush.
    \item A cup of water is falling towards the ground on its side.
\end{itemize}
\normalfont
}

\section{Additional Qualitative Results}
\label{supp_sec:qualitative}

\begin{figure}[t]
  \centering
  \includegraphics[width=0.99\linewidth]{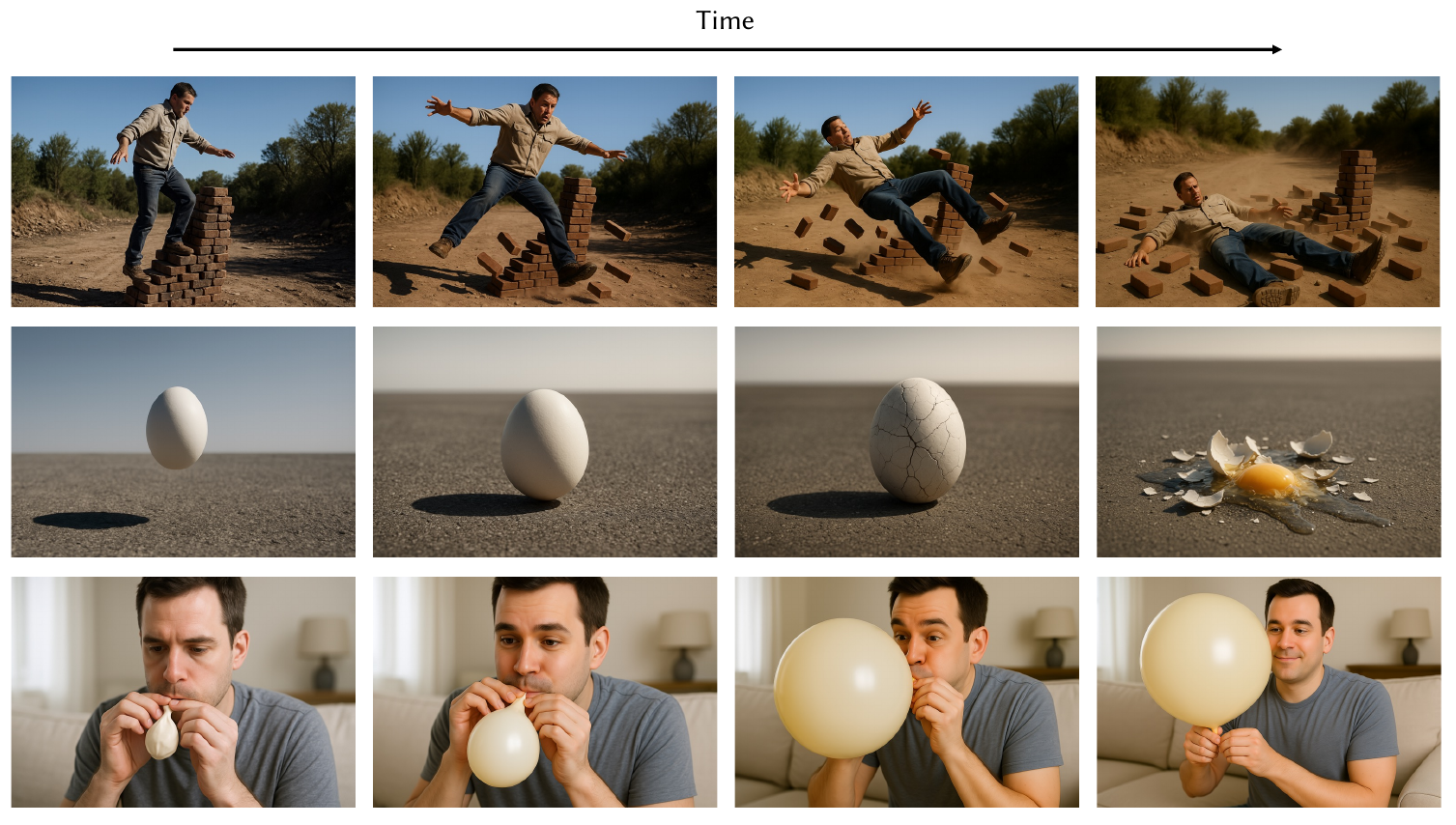}
  \caption{{\textbf{GPT Keyframe Limitations.}
  Qualitative examples showing the accumulated saturation and smoothness artifacts produced by \texttt{gpt-image-1} during iterative keyframe generation. 
  As each generated image is recursively used as part of the input for the next step, slight over-saturation and over-smoothing compound over time, leading to slight color shifts (\eg, yellow cast) and reduced photorealism in later frames.}}
  \label{fig:fig_limitation}
\end{figure}

We present additional qualitative examples illustrating the saturation limitations of Visual Thought Generation in Figure~\ref{fig:fig_limitation}.

\begin{figure*}[t]
  \centering
  \includegraphics[width=\linewidth]{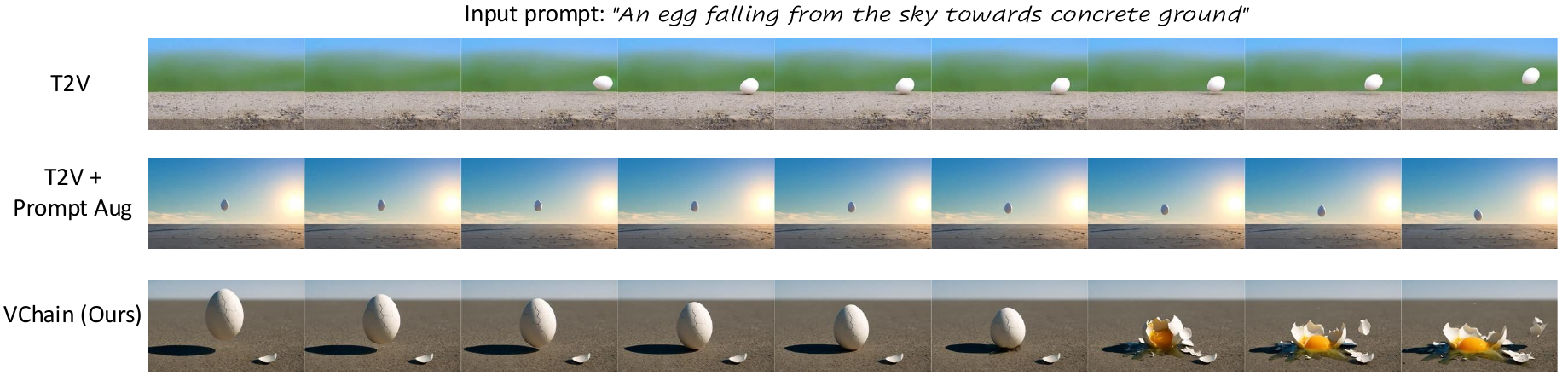}
  \caption{\textbf{Additional Qualitative Comparisons - Egg Fall.}}
  \label{fig:fig_paper_comparison_baseline_egg_ground}
  \vspace{20pt}
\end{figure*}

\begin{figure*}[t]
  \centering
  \includegraphics[width=\linewidth]{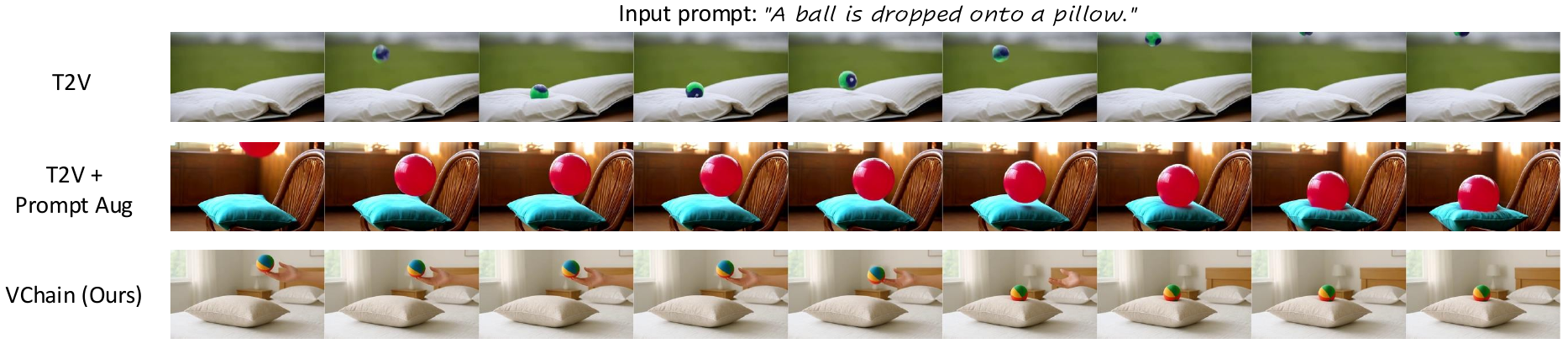}
  \caption{\textbf{Additional Qualitative Comparisons - Pillow.}}
  \label{fig:fig_paper_comparison_baseline_pillow}
  \vspace{20pt}

\end{figure*}

\begin{figure*}[t]
  \centering
  \includegraphics[width=\linewidth]{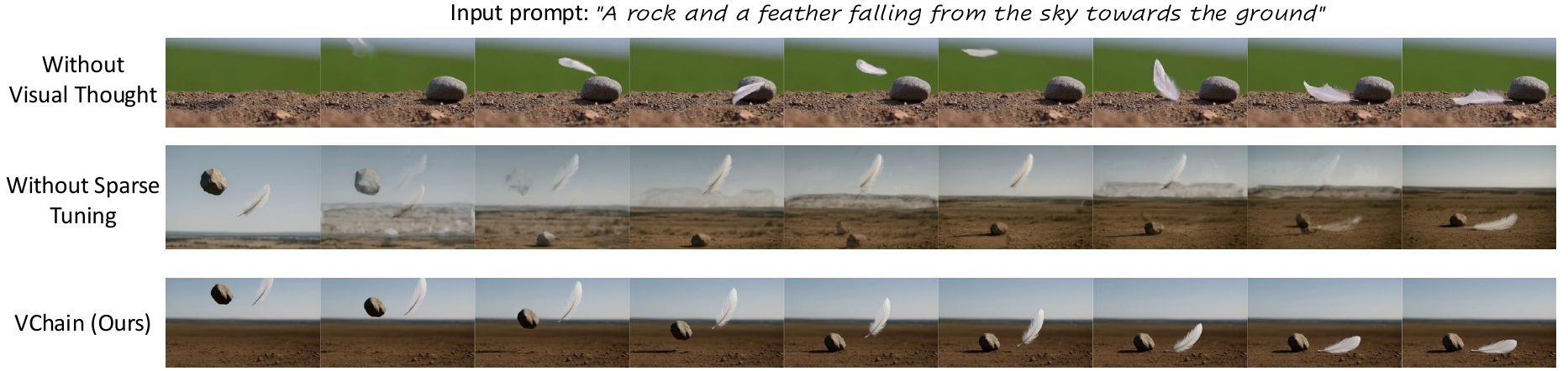}
  \caption{\textbf{Additional Qualitative Comparisons - Rocket Feather.}}
  \label{fig:fig_paper_comparison_ablation_rock_feather}
    \vspace{20pt}

\end{figure*}

\begin{figure*}[t]
  \centering
  \includegraphics[width=\linewidth]{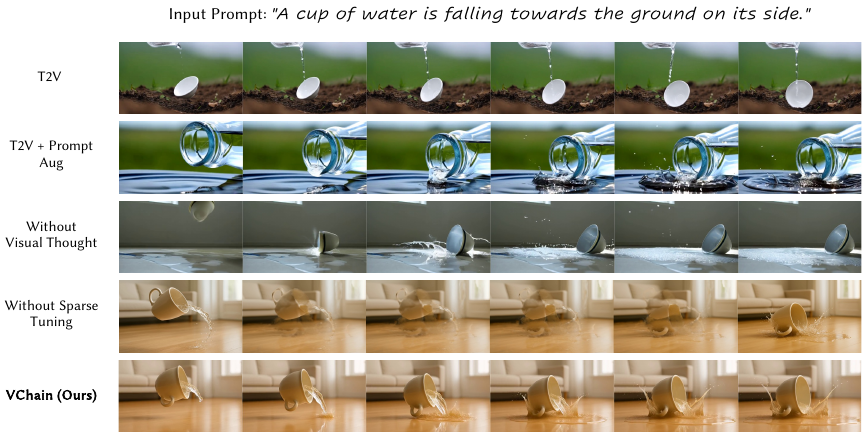}
  \caption{\textbf{Additional Qualitative Comparisons - Cup.}}
    \label{fig_supp_1}
  \vspace{10pt}

\end{figure*}

\begin{figure*}[t]
  \centering
  \includegraphics[width=\linewidth]{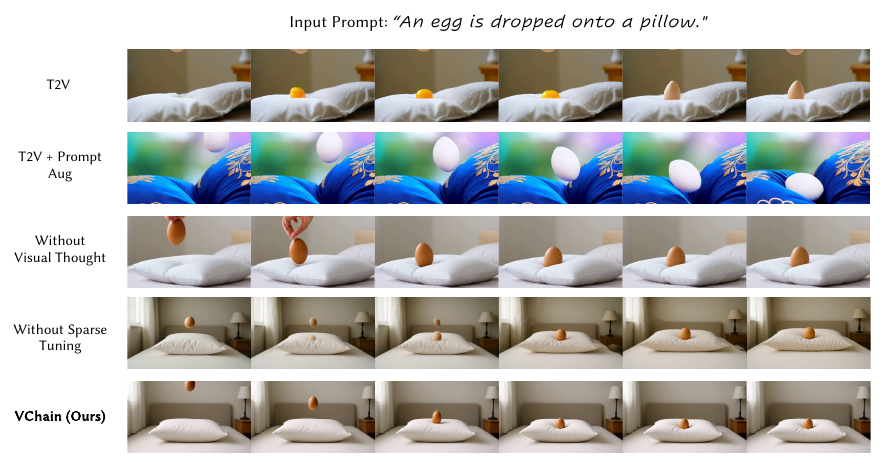}
  \caption{\textbf{Additional Qualitative Comparisons - Egg Pillow.}}
    \label{fig_supp_2}
  \vspace{20pt}

\end{figure*}

\begin{figure*}[t]
  \centering
  \includegraphics[width=\linewidth]{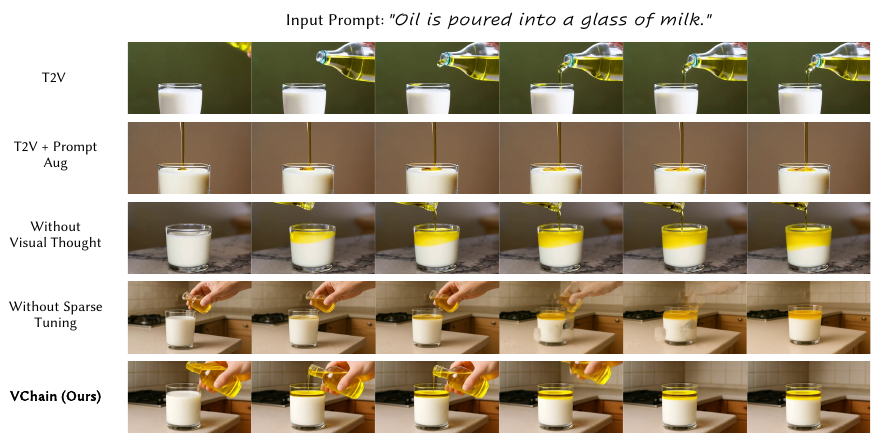}
  \caption{\textbf{Additional Qualitative Comparisons - Oil Milk.}}
    \label{fig_supp_3}
  \vspace{10pt}

\end{figure*}

\begin{figure*}[t]
  \centering
  \includegraphics[width=\linewidth]{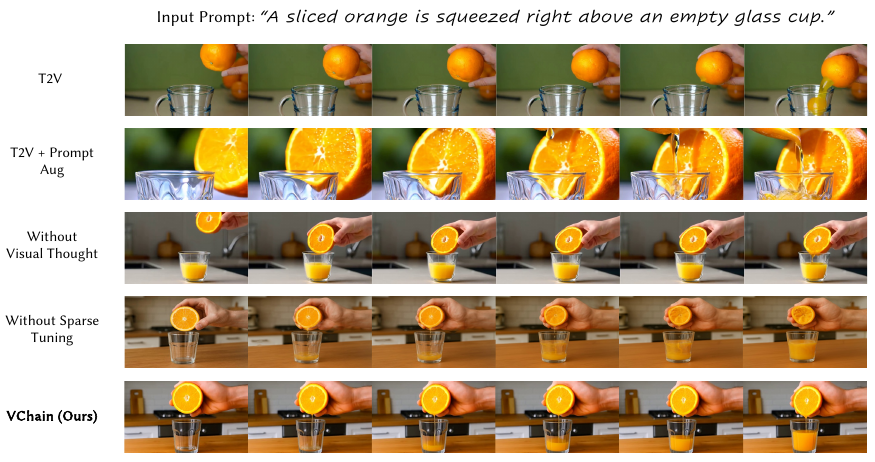}
  \caption{\textbf{Additional Qualitative Comparisons - Orange.}}
  \label{fig_supp_4}
    \vspace{20pt}

\end{figure*}

Further qualitative comparisons are shown in Figures~\ref{fig:fig_paper_comparison_baseline_egg_ground}, \ref{fig:fig_paper_comparison_baseline_pillow}, \ref{fig:fig_paper_comparison_ablation_rock_feather}, \ref{fig_supp_1}, \ref{fig_supp_2}, \ref{fig_supp_3}, and~\ref{fig_supp_4}.

\balance              

\clearpage

\begin{figure*}[h] 
      \centering
    \begin{tcolorbox}[
        title={First Frame System Message},
        fonttitle=\ttfamily\bfseries,
        colframe=black!70!white,
        colback=black!5!white,
        fontupper=\small\ttfamily,
        width=\textwidth,
        before upper={\setlength{\parskip}{8pt}\setlength{\parindent}{0pt}}
    ]
    
    {\ttfamily
    
    You are given a text prompt, which describes a video. You are to perform the following tasks:
    
    1. Infer the objects/people/elements present in the scene, the perspective of the camera, the spatial relationship between the objects in the scene as well as details not explicitly mentioned in the text prompt.
    
    2. Create a detailed, movie-like description of the scene that evokes visuals with strong detail and composition cues. This is the Context Frame. It should clearly describe the objects/people/elements present in the scene, the perspective of the camera, and the spatial relationships between the objects in it as well as the details not explicitly mentioned in the text prompt.
    
    The context frame must depict the initial state of the scene, before any action occurs, and must not foreshadow the input prompt. Given an input prompt "A man throws a ball", the context frame should depict a man holding a ball at his side, not in mid-throw. Given an input prompt "A man squeezes a ball in his hand", the context frame should depict a man gently holding a ball in his palm, not squeezing it yet. Given an input prompt "A dolphin emerges from the water", the context frame should depict a calm ocean. The dolphin should not be visible yet. The context frame should be written as if it is depicting an image, not a video. Hence, it should not foreshadow what will happen next.
    
    3. Create a concise version of the context frame. This should be a short, one-sentence description of the context frame. 
    
    4. Infer a sequence of consequences/changes  from the text prompt, even if it is not explicitly mentioned. 
    Use assertive languange to clearly describe the changes in appearance, shape, color, size, and position that may occur as a result.

    Example:
    
    --------------
    
    Input Prompt: "A cat pushes a glass of water off a table."
    
    Thoughts: In order for the cat to tip the glass off the table, the cat is sitting on the table next to the glass of water. In order for the glass of water to fall off the table, it should be placed precariously on the edge of the table. A side view perspective would capture the table, the cat, the glass of water in one frame.  
    
    Context Frame: A side view of a sleek tabby cat sitting upright on a wooden table in a kitchen. The glass of water is placed precariously at the very edge of the table. The cat gazes intently at the glass, its tail curled around its body. The camera is at mid-height, framing the cat, table, glass, and floor clearly in the shot.
    
    Concise Prompt: A cat sits next to a glass of water on a table.
    
    Consequences: The cat will touch the glass of water, causing it to tip over the edge of the table and fall towards the ground. The glass of water will touch the ground and shatter as a result. The water will spill everywhere and glass shards will be on the floor.
    
    --------------
    Additional Examples
    --------------

    }
    
    \end{tcolorbox}
    \caption{\textbf{First Frame System Message.}}
      \label{fig:first_frame_system_message}
\end{figure*}

\begin{figure*}[h] 
      \centering
    \begin{tcolorbox}[
        title={Next Frame System Message},
        fonttitle=\ttfamily\bfseries,
        colframe=black!70!white,
        colback=black!5!white,
        fontupper=\small\ttfamily,
        width=\textwidth,
        before upper={\setlength{\parskip}{8pt}\setlength{\parindent}{0pt}}
    ]
    
    {\ttfamily
     You are given an input prompt, which describes a video. You are also given a sequence of keyframes (1 or more keyframes), meant to depict key moments of the video. You are also given a hint, describing what happens throughout the video. You are to predict the next keyframe in the sequence. Use precise language to clearly describe the changes that may occur in this next key. You must predict what may happen within the next 5 seconds of the video. Hence, do not predict too far into the future.
    
    A keyframe is a still image that captures either the start, peak/intermediate stage, the end, or the consequence of an event. 
    The next predicted keyframe MUST ONLY depict either the start, peak, end, or consequence of an event and NEVER a combination of them. 
    
    e.g. The key moments of kicking a ball into a goal are (1) The moment the foot makes contact with the ball. The ball should not have moved at this point., (2) The moment the ball is inside the goal.
    e.g. The key moments of ice melting are (1) When the ice is fully solid (2) The moment the ice cube is half melted (3) The moment the ice cube is completely melted with a large puddle of water.
    e.g. The key moments of a glass of water falling off a table are (1) The moment the glass of water is on the edge of the table, (2) The moment the glass of water is falling midair towards the ground (3) The moment the glass of water makes contact with the ground but is still in one piece. (4) The moment the glass of water shatters on the ground and the water spills everywhere.
    
    If the next key moment involves contact between two objects, then the next keyframe must depict the moment of contact. The objects must be touching in the next predicted key rame description. Your caption for the next keyframe should not use comparative language to describe a relative change in position, distance, or size (e.g. towards, away from). Instead, it should describe the absolute position, distance, or size of the objects involved. If possible, use spatial prepositions to clarify the relationship between objects (e.g. inside of, on top of, and below).
    The next keyframe should describe the image as if it not in motion. Hence, avoid using phrases like 'about to', 'going to'. Describe the next keyframe as if it is a still image.
    
    Finally, return 'True' if the next predicted keyframe is the last frame of this video, otherwise, return 'False'. If nothing significant happens after the next predicted keyframe, return 'True'.

    Example:
    
    --------------
    
    Input Prompt: "A cat pushes a glass of water off a table"
    
    Hint: The glass of water will fall off and shatter on the floor. The water will spill everywhere and glass shards will be on the floor.
    
    Keyframe 1: [A cat is sitting on the table, and the glass of water is on the edge of the table.]
    
    Next Predicted Keyframe: The cat's paw is touching the glass of water sitting on the table.
    
    Last Frame: False
    
    --------------
    Additional Examples
    --------------
    
    }
    
    \end{tcolorbox}
    \caption{\textbf{Next Frame System Message.}}
      \label{fig:next_frame_system_message}
\end{figure*}

\lstdefinelanguage{json}{
  basicstyle=\ttfamily\footnotesize,
  showstringspaces=false,
  breaklines=true,
  stringstyle=\color{blue!30!black},
  morestring=[b]",
  literate=
   *{0}{{{\color{blue}0}}}{1}
    {1}{{{\color{blue}1}}}{1}
    {2}{{{\color{blue}2}}}{1}
    {3}{{{\color{blue}3}}}{1}
    {4}{{{\color{blue}4}}}{1}
    {5}{{{\color{blue}5}}}{1}
    {6}{{{\color{blue}6}}}{1}
    {7}{{{\color{blue}7}}}{1}
    {8}{{{\color{blue}8}}}{1}
    {9}{{{\color{blue}9}}}{1}
}
\begin{figure*}[h] 
      \centering
    \begin{tcolorbox}[
        title={Reasoning Output Example},
        fonttitle=\ttfamily\bfseries,
        colframe=black!70!blue,
        colback=blue!1!white,
        fontupper=\small\ttfamily,
        width=\textwidth,
        before upper={\setlength{\parskip}{8pt}\setlength{\parindent}{0pt}}
    ]
    \begin{lstlisting}[language=json]

    "sulfuric_acid": {
        "input_prompt": "Concentrated sulfuric acid is poured onto a wooden table.",
        "thoughts": "The scene involves a wooden table, likely in a laboratory or workshop setting, where concentrated sulfuric acid is about to be poured. The acid is typically stored in a glass or plastic container, and the person pouring it might be wearing protective gear such as gloves and goggles. The camera should capture a side view to show the table, the container of acid, and the person pouring it. The table is initially dry and intact, with visible wood grain.",
        "consequences": "As the sulfuric acid is poured onto the wooden table, it will react with the wood, causing it to char and emit smoke. The wood will darken and potentially start to disintegrate where the acid makes contact, creating a burnt, uneven surface. The reaction may produce heat and release fumes, necessitating proper ventilation and safety precautions.",
        "context_frame": "In a well-lit laboratory, a sturdy wooden table stands at the center of the scene, its surface smooth and polished, with visible wood grain patterns. A person, wearing protective gloves and goggles, stands beside the table, holding a glass container filled with concentrated sulfuric acid. The container is tilted slightly, poised to pour. The camera captures a side view, framing the table, the container, and the person, highlighting the contrast between the clear, viscous liquid and the warm tones of the wood.",
        "concise_prompt": "A person stands beside a wooden table, holding a container of concentrated sulfuric acid.",
        "key_frames": [
            "The area where the concentrated sulfuric acid makes contact with the wooden table starts to darken and emit smoke. The wood grain appears charred and blackened, with visible smoke rising from the surface. The edges of the darkened area are irregular, indicating the beginning of disintegration.",
            "The concentrated sulfuric acid has been poured onto the wooden table. A small, blackened, and charred area is visible on the table where the acid has made contact. Smoke is rising from the reaction site, and the wood grain around the area has started to darken and disintegrate slightly, illustrating the corrosive impact of the acid.",
            "The concentrated sulfuric acid creates a deep, blackened mark on the wooden table where it has been poured. The wood is significantly charred with smoke wafting upwards, forming a small plume. The surrounding area of the wood appears darker, with slight disintegration at the center of the spill, indicating intense chemical reaction.",
            "The concentrated sulfuric acid has reacted with the wooden table, and the area of contact has expanded. The wood appears darkened and severely burnt, with visible smoke and fumes rising prominently into the air. The wooden surface is visibly damaged, with large burnt patches and disintegrated material, showing an uneven and charred texture. The person remains focused on observing the reaction, and the container of acid is still slightly tilted above the table."
        ]
    }

    \end{lstlisting}
    \end{tcolorbox}
    \caption{\textbf{Reasoning Output Example.}}
      \label{fig:reasoning_output_example}
\end{figure*}

\lstdefinelanguage{txt}{
  showstringspaces=false,
  breaklines=true,
  basicstyle=\ttfamily\footnotesize,
  columns=fullflexible
}
\begin{figure*}[t]
      \centering
      \begin{tcolorbox}[
        title={CSV Output Example},
        fonttitle=\ttfamily\bfseries,
        colframe=black!70!green,
        colback=green!1!white,
        fontupper=\small\ttfamily,
        width=\textwidth,
        before upper={\setlength{\parskip}{8pt}\setlength{\parindent}{0pt}}
      ]
    \begin{lstlisting}[language=txt]
    "file_name","text"
    "sulfuric_acid_0.png", "A person stands beside a wooden table, holding a container of concentrated sulfuric acid."
    "sulfuric_acid_1.png","The area where the concentrated sulfuric acid makes contact with the wooden table starts to darken and emit smoke. The wood grain appears charred and blackened, with visible smoke rising from the surface. The edges of the darkened area are irregular, indicating the beginning of disintegration."
    "sulfuric_acid_2.png","The concentrated sulfuric acid has been poured onto the wooden table. A small, blackened, and charred area is visible on the table where the acid has made contact. Smoke is rising from the reaction site, and the wood grain around the area has started to darken and disintegrate slightly, illustrating the corrosive impact of the acid."
    "sulfuric_acid_3.png","The concentrated sulfuric acid creates a deep, blackened mark on the wooden table where it has been poured. The wood is significantly charred with smoke wafting upwards, forming a small plume. The surrounding area of the wood appears darker, with slight disintegration at the center of the spill, indicating intense chemical reaction."
    "sulfuric_acid_4.png", "The concentrated sulfuric acid has reacted with the wooden table, and the area of contact has expanded. The wood appears darkened and severely burnt, with visible smoke and fumes rising prominently into the air. The wooden surface is visibly damaged, with large burnt patches and disintegrated material, showing an uneven and charred texture. The person remains focused on observing the reaction, and the container of acid is still slightly tilted above the table."

    \end{lstlisting}
      \end{tcolorbox}
      \caption{\textbf{CSV Output Example.}}
      \label{fig:csv_output_example}
\end{figure*}

\begin{table*}[h]
  \caption{\textbf{Time Cost Breakdown.}}
  \label{tab:time_cost}
\resizebox{\textwidth}{!}{
    \begin{tabular}{l|lll}
    \toprule
    \textbf{Breakdown} & \textbf{\Centerstack{Time Cost}} & \textbf{\Centerstack{Comments}}   \\    \midrule 
    \textbf{Visual Thought Reasoning} & 3 min 3 sec &  \\ \hline
    initial reasoning & 14 sec & API: \textit{gpt-4o} chat completions, called once for every sequence, CPU \\
    image generation & 1 min 7 sec & API: \texttt{gpt-image-1} generate, called once for every sequence, CPU \\
    image perception & 16 sec & API: gpt-4o vqa, called 2.5 times (Averaged across 35 sequences), CPU \\
    image editing &  1 min 26 sec & API: \texttt{gpt-image-1} edit, called  2.5 times (Averaged across 35 sequences), CPU \\ \toprule\hline
    \textbf{Sparse Inference-Time Visual-State Adaptation} & 5 min 36 sec & Wan2.1-T2V-1.3B, 480$\times$832, 81 frames, NVIDIA A100 GPU \\ \hline
    pre-process visual thoughts for fine-tuning & 30 sec &  \\
    load model & 6 sec &  \\
    fine-tuning &  5 min & including checkpoint saving \\ \toprule\hline
    \textbf{Sparse Inference-Time Visual-State Adaptation} & 6 min 56 sec & Wan2.1-T2V-14B, 480$\times$832, 81 frames, NVIDIA A100 GPU \\ \hline
    pre-process visual thoughts for fine-tuning & 30 sec &  \\
    load model & 20 sec &  \\
    fine-tuning &  6 min 6 sec & including checkpoint saving \\ \toprule\hline
    \textbf{Video Sampling} & 3 min 9 sec & Wan2.1-T2V-1.3B, 480$\times$832, 81 frames, NVIDIA A100 GPU \\ \hline
    model loading & 14 sec & could save time by not saving then re-loading checkpoint upon tuning \\
    sampling & 2 min 46 sec &  \\
    VAE decoding \& video saving &  9 sec & \\  \toprule\hline
    \textbf{Video Sampling} & 14 min 48 sec & Wan2.1-T2V-14B, 480$\times$832, 81 frames, NVIDIA A100 GPU \\ \hline
    model loading & 33 sec & could save time by not saving then re-loading checkpoint upon tuning \\
    sampling & 14 min 06 sec &  \\
    VAE decoding \& video saving &  9 sec & \\

    \bottomrule
    \end{tabular}
}
\end{table*}

\begin{figure*}[t]
  \centering
  \includegraphics[width=\linewidth]{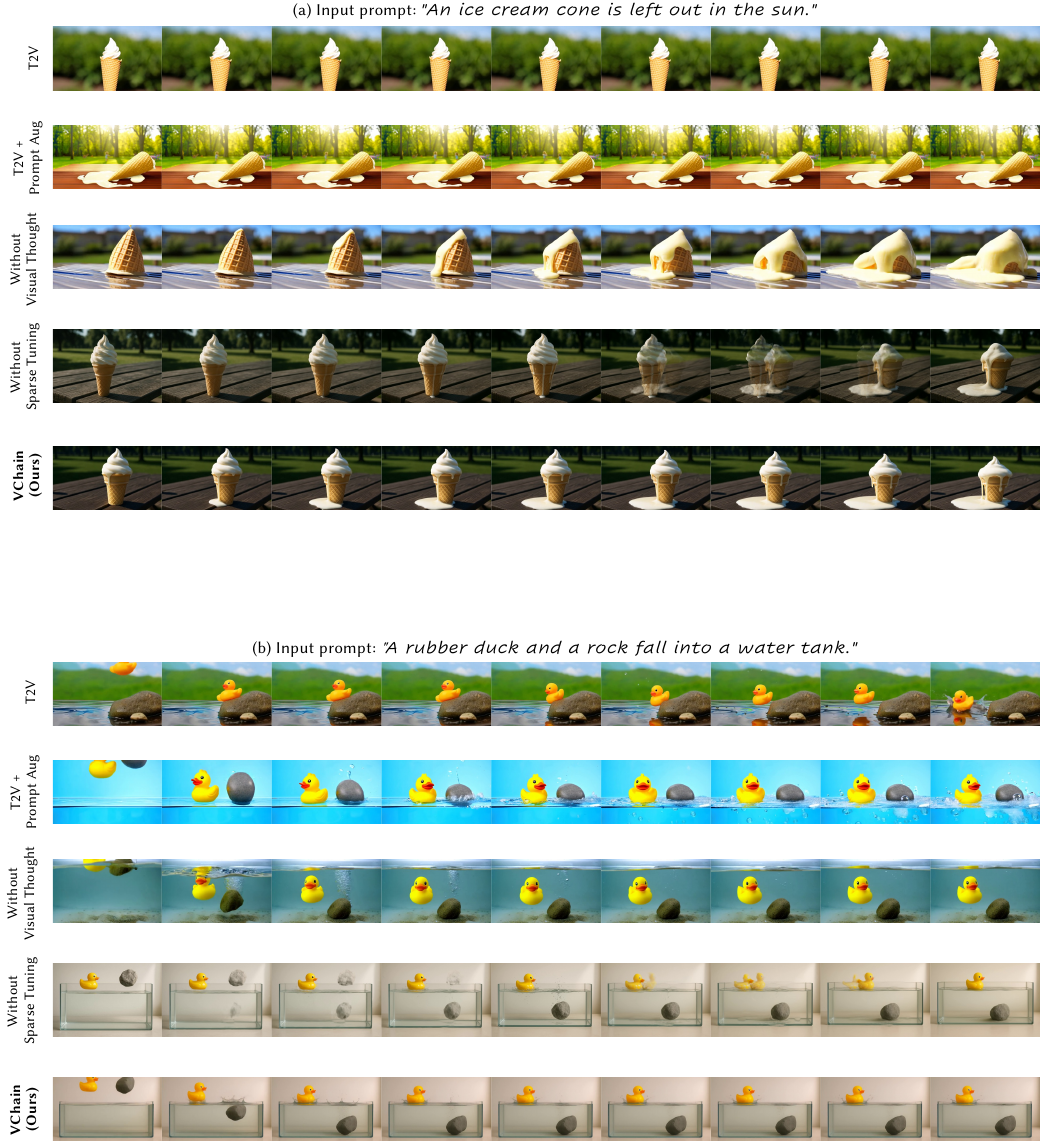}
  \caption{\textbf{More Qualitative Comparisons.} }
  \label{fig:fig_paper_comparison_5baselines}
\end{figure*}

\clearpage

\begin{figure*}[t]
  \centering
  \includegraphics[width=\linewidth]{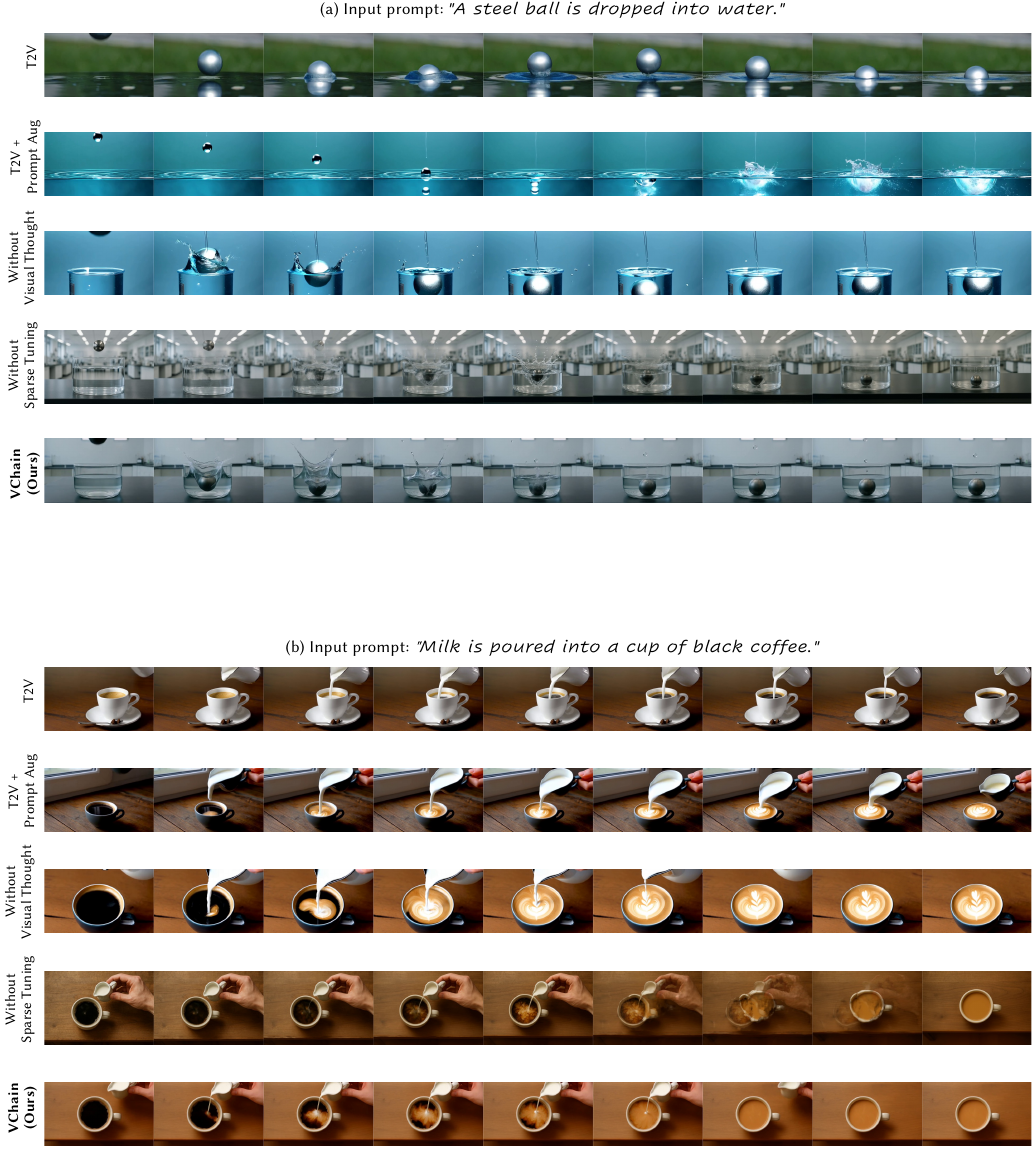}
  \caption{\textbf{More Qualitative Comparisons.}}
  \label{fig:fig_paper_comparison_5baselines_steelball_coffeemilk}
\end{figure*}

\clearpage


\clearpage

\clearpage

\clearpage

\section{Discussion: Evolution of Reasoning Paradigms}
\label{sec:reasoning_discussion}

To better situate our framework, we discuss the evolution of reasoning paradigms and their application in the context of video generation and general generative tasks.

\subsection{From Textual to Visual Reasoning in the Era of Large Foundation Model} 
Before the era of Large Foundation Models (LFMs), reasoning has long been studied as a core paradigm for solving complex problems (e.g., symbolic reasoning, logical inference, and analogical reasoning), and has been widely applied in both NLP and vision tasks to handle structured and multi-step decision making~\cite{jiao2022merit,huang2021diagnostic}. 
With the emergence of LFMs, new reasoning paradigms have arisen based on their strong emergent capabilities. Representative textual reasoning approaches include \textit{explicit reasoning} methods such as Chain-of-Thought (CoT)~\cite{wei2022chain} and Tree-of-Thought (ToT)~\cite{yao2023tree}, as well as \textit{implicit reasoning} that performs multi-step computation within hidden representations~\cite{hao2024training}. 
Recent advances, such as DeepSeek-R1~\cite{guo2025deepseek}, OpenAI-o1~\cite{openai2025reasoning}, and Mythos~\cite{anthropic2026mythos}, demonstrate that step-by-step reasoning significantly enhances performance, achieving strong or even superhuman results across a wide range of domains, including mathematics~\cite{team2025kimi}, programming~\cite{hui2024qwen2}, software engineering~\cite{wei2025swe},  personalization~\cite{lin2026bringing, lin2026verifiable}, and safety~\cite{jiang2025safechain,zhang2025safety}. 

Despite these successes, directly transferring such reasoning paradigms from the linguistic space to the visual space remains challenging. 
In particular, applying textual reasoning (\textit{e.g.}, CoT) to video generation often leads to a fundamental \textbf{semantic gap}: while a model may correctly describe a causal sequence in natural language, the underlying video generator may lack the physical, spatial, and temporal priors required to faithfully execute these transitions. This gap highlights the limitation of purely language-based reasoning when dealing with inherently visual and dynamic processes. To address this limitation, our work introduces \textbf{Chain of Visual Thoughts}, which shifts reasoning from textual or latent linguistic space to explicitly grounded visual states, enabling more faithful video generation.

\subsection{Taxonomy of Reasoning in Video Generation}

We categorize the existing and emerging reasoning paradigms in video generation into four distinct levels:

\noindent\textbf{Implicit Reasoning (End-to-End):} Most foundation models~\cite{wan2025wan, wiedemer2025video} rely on vast datasets to implicitly learn world physics in a data-driven way. While they produce visually fluid motion, they could suffer from ``causal hallucinations'' in multi-step scenarios, such as objects violating gravity or failing to reflect logical consequences.

\noindent\textbf{Text-Guided Planning (Two-Stage):} This decoupled paradigm uses LLMs as high-level planners to generate intermediate scripts or layouts (\textit{e.g.},~\cite{lian2023lmd, lian2023lvd, li2023gligen, qu2023layoutllm, wang2026mavis}). It can largely improve global structure, layout, and plots, but it could struggle with fine-grained visual-state transitions that are difficult to describe purely in text or bounding boxes.

\noindent\textbf{MLLM-aided Reasoning (Feature-Level):} This paradigm utilizes MLLMs to provide latent features, either generated or queried, to aid a generative decoder~\cite{pan2025transfer, lin2025exploring}. These methods typically require dense retraining and structural modifications to align multimodal embeddings with the generator.

\noindent\textbf{Visual Thought Reasoning (VChain):} We introduce a ``Chain-of-\textit{Visual}-Thought'' paradigm that externalizes reasoning as sparse visual keyframes. This allows the video generator to adapt its weights at inference time to specific, physically grounded visual states via \textit{Sparse Inference-Time Visual-State Adaptation}.

\subsection{Key Messages and Future Outlook}
Based on our findings, we summarize two key takeaways regarding the future of reasoning in video generation:
\begin{enumerate}
    \item \textbf{The Necessity of Direct Visual CoT:} Our experiments reveal that for video generation, CoT must be \textit{directly visual} rather than purely textual. While textual reasoning provides a logical blueprint, only visual thoughts provide the explicit spatial and material constraints (\textit{e.g.}, the exact details of a splash or the buoyancy of an object) necessary to override the incorrect physical priors of a generator. Visual thoughts act as a bridge that translates symbolic logic into pixel-level consistency.
    \item \textbf{The Reasoner-Renderer Paradigm:} \texttt{VChain} establishes a modular paradigm where a powerful \textit{Reasoner} (e.g., an MLLM like GPT-4o and Gemini) guides a specialized \textit{Renderer} (e.g., a video diffusion transformer). Currently, MLLMs exhibit superior semantic and symbolic reasoning capabilities compared to native video models. As we prepare this final version, we observe emerging research suggesting that advanced, large-scale video generators are beginning to develop potential reasoning abilities, particularly in \textit{spatial reasoning}~\cite{wiedemer2025video, wang2026demystifing}. This suggests a future where the boundary between \textit{Reasoner} and \textit{Renderer} may blur, though the structured guidance provided by frameworks like \texttt{VChain} remains essential in reasoning for complex, multi-step causal consistency. As long as the reasoning ability of state-of-the-art MLLMs continues to outpace that of video generators (a gap that might persist for a while), VChain remains an effective way to transfer such reasoning into the generation process.
\end{enumerate}